\documentclass[sn-mathphys-num]{sn-jnl}

\usepackage{color}
\usepackage{graphicx}
\usepackage{times}
\usepackage{amsmath,amssymb,amsfonts}
\usepackage{bm}
\usepackage{url}
\usepackage{subcaption}
\usepackage{multirow}

\begin{document}

\title{Clustering Change Sign Detection by Fusing Mixture Complexity}


\author*[1]{\fnm{Kento} \sur{Urano}}\email{uranokento38@gmail.com}

\author[1]{\fnm{Ryo} \sur{Yuki}}\email{ling.jiecheng.rd@gmail.com}

\author[1]{\fnm{Kenji} \sur{Yamanishi}}\email{yamanishi@g.ecc.u-tokyo.ac.jp}


\affil*[1]{\orgdiv{School of Information Science and Technology}, \orgname{The University of Tokyo}, \orgaddress{\street{7-3-1 Hongo, Bunkyoku}, \city{Tokyo}, \postcode{113-8656}, \country{Japan}}}




\abstract{
This paper proposes an early detection method for cluster structural changes.
Cluster structure refers to discrete structural characteristics, such as the number of clusters, when data are represented using finite mixture models, such as Gaussian mixture models. We focused on scenarios in which the cluster structure gradually changed over time.
For finite mixture models, the concept of mixture complexity (MC) measures the continuous cluster size by considering the cluster proportion bias and overlap between clusters.
In this paper, we propose MC fusion as an extension of MC to handle situations in which multiple mixture numbers are possible in a finite mixture model.
By incorporating the fusion of multiple models, our approach accurately captured the cluster structure during transitional periods of gradual change. Moreover, we introduce a method for detecting changes in the cluster structure by examining the transition of MC fusion. We demonstrate the effectiveness of our method through empirical analysis using both artificial and real-world datasets.
}

\keywords{
clustering, change detection, information theory, mutual information}

\maketitle

\section{Introduction}
\subsection{Motivation}
This study focused on detecting changes in the underlying cluster structure of time-series data and identifying the signs of these changes.
Cluster structure refers to discrete structural characteristics, such as the number of clusters, when data are represented using finite mixture models, such as Gaussian mixture models. Furthermore, we investigated the differences in cluster structure by examining the cluster proportions and overlaps between clusters.
We consider scenarios in which the cluster structure changes over time.
The detection of such changes is crucial because they correspond to significant real-world events. For example, in the context of modeling consumer purchase data using a Gaussian mixture model, changes in the number of clusters indicate the emergence or disappearance of purchase patterns, which reflect shifts in market trends~\cite{HY12}.

\begin{figure}
	\centering
	\includegraphics[width=\columnwidth]{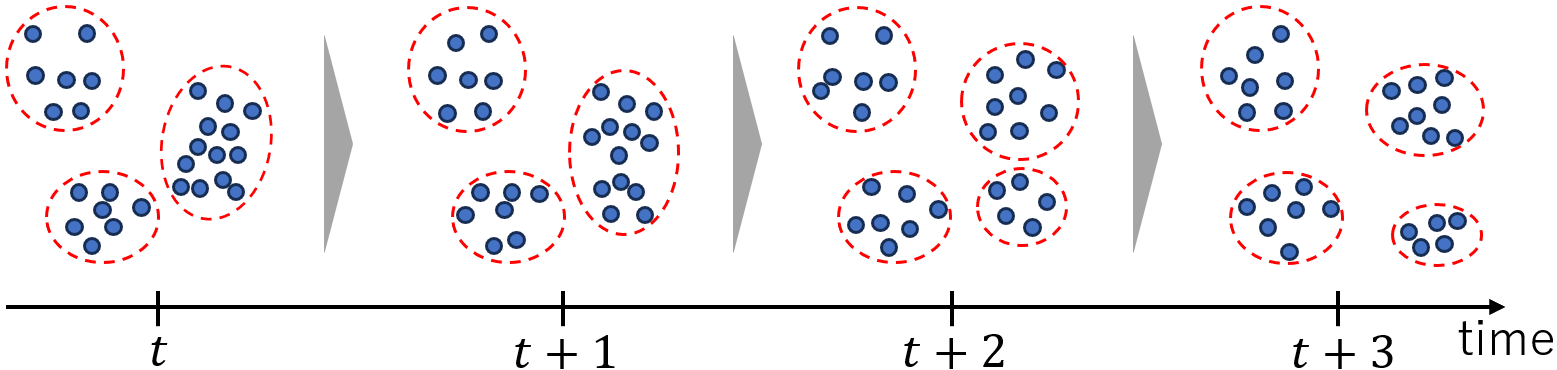}
	\caption{Transition of the number of clusters gradually changing from 3 to 4.}
	\label{intro_gradual}
\end{figure}

The field of structural change detection has been extensively investigated ~\cite{HW98,K03,YF18,YM07}. However, many existing approaches assume sudden changes due to the discrete nature of the underlying structures. It is natural to consider that structural changes occur gradually, as shown in Fig. ~\ref{intro_gradual}. The concept of structural change sign detection was proposed to detect gradual structural changes at an early stage.

\begin{figure}
	\centering
	\includegraphics[width=\columnwidth]{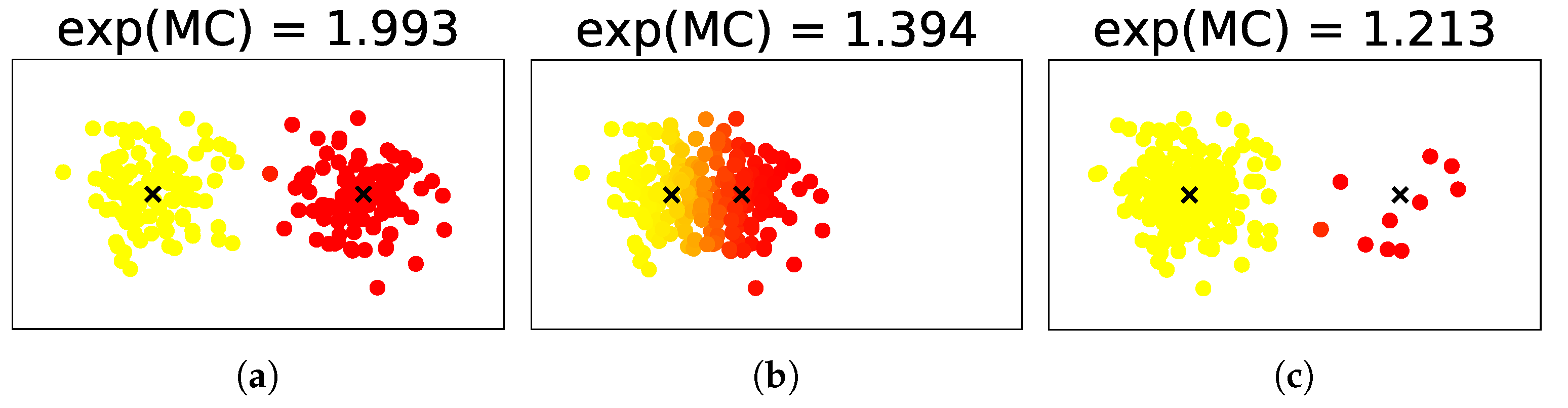}
	\caption{Example of MC for a Gaussian Mixture Model with a mixture size of 2~\cite{KY22}.}
	\label{mc_intro}
\end{figure}

An applicable concept for detecting signs of changes in the cluster structure is mixture complexity (MC)~\cite{KY22}. MC is a measure of the continuous cluster size for finite mixture models that considers the bias in cluster proportions and the overlap between clusters. For instance, consider the data generated from the Gaussian mixture model depicted in Fig. ~\ref{mc_intro}. Although all three cases had the same number of components (two), the cluster structures differed. In case (a), the two clusters were separate, with no bias in the cluster proportions. Hence, it is reasonable to assume that the cluster size is 2. However, in case (b), the means of the two clusters were close and overlapped. In case (c), the data are concentrated in one cluster with bias. Therefore, asserting that the cluster size is 2 for cases (b) and (c) is problematic. By utilizing MC, which provides a continuous extension of cluster size, the corresponding values for these examples were 1.99, 1.39, and 1.21.

MC is a measure that calculates the mutual information between the observed variable $X$ and the latent variable $Z \in \{1, \ldots, k\}$, which indicates the cluster from which the observed variable $X$ is generated in a finite mixture model with $k$ mixture components.
MC itself holds theoretical value for evaluating the cluster structure in finite mixture models. However, MC is based on the assumption that the distribution is known, meaning that the number of mixtures and parameters of the distributions in the finite mixture model are fixed and known. When applying MC to sign detection, where the distribution is unknown, the calculated MC values may not accurately represent the cluster sizes. In sign detection, the MC values are calculated using the estimated number of components, making the MC dependent on the estimated number of components at each time step. This dependency poses a challenge as it can lead to an inaccurate capture of the cluster size, especially when the number of components is underestimated. Consequently, the early detection of an increase in the number of clusters using the structural change sign detection method based on MC is particularly challenging.

Other studies on structural change sign detection include Descriptive Dimensionality (Ddim)~\cite{YH23}, which quantifies the dimensionality of models based on the Minimum Description Length (MDL) principle~\cite{R78,Y23} under the assumption of stochastic fusion of multiple models. Another approach is structural entropy (SE)~\cite{HY18}, which measures the uncertainty associated with model changes. However, Ddim suffers from the limitation that the dimensionality of finite mixture models, which mix together, is solely determined by the respective number of components, neglecting the differences in structure caused by overlap and bias. Consequently, it fails to effectively capture gradual changes such as the emergence or disappearance of clusters, making early detection difficult. The same limitation applies to the SE.

To address these challenges, this study proposes an extension of MC called MC fusion, which accommodates cases in which multiple components are possible in finite mixture models. By extending MC in this manner, we preserve its advantages in evaluating cluster structures while enabling the early detection of increases in the number of clusters, which is a challenge with traditional MC. The effectiveness of the cluster structure change sign detection method using MC fusion is demonstrated through experiments conducted on artificial and real-world datasets.

\subsection{Related Work}
In the field of model selection, various information criteria, such as AIC~\cite{A74}, BIC~\cite{S78}, and MDL~\cite{R78} have been applied to different scenarios. The MDL principle~\cite{R78,Y23}, which minimizes the normalized maximum likelihood (NML) code length, has been extensively studied for model selection owing to its optimality in terms of Shtarkov's minimax regret~\cite{S87} and its fast convergence in stochastic PAC learning scenarios ~\cite{Y92}.

Although model selection has primarily focused on stationary probabilistic models, research has also investigated methodologies for situations in which models change over time. Model change detection methods aim to identify changes in the underlying model structure of time-series data rather than changes in the distribution parameters. These changes may involve alterations in the number of distribution parameters or clusters. In the context of model change detection based on the MDL principle, progress has been made using the Dynamic Model Selection (DMS) algorithm~\cite{YM05, YM07}. The DMS algorithm outputs a sequence of models that minimize code length, and Hirai and Yamanishi proposed the Sequential Dynamic Model Selection (SDMS) algorithm, which applies DMS to sequentially obtained data~\cite{HY12}. Yamanishi and Fukushima extended the MDL change statistic~\cite{YM16}, originally proposed for parameter change detection, to model change detection and justified the DMS in the context of hypothesis testing through a new theoretical analysis~\cite{YF18}. Model change detection was also explored for other scenarios. Herbster and Warmuth proposed a method that tracks the best expert by iteratively updating the weights of the model candidates~\cite{HW98}. Erven et al. defined the concept of switching distributions and presented an algorithm that selects a sequence of models by maximizing their posterior probabilities ~\cite{EGR12}. Kleinberg proposed a burst-detection algorithm that estimates the transitions of latent states~\cite{K03}. Huang et al. developed a method for detecting changes in the rate of detected changes, known as volatility shift~\cite{HKDP14}.

The aforementioned research on model change detection primarily focused on discrete changes in models, assuming that model changes occur abruptly. However, gradual changes in the model are often more prevalent. Consequently, recent studies have concentrated on developing model change sign detection methods capable of detecting progressive model changes at earlier stages. Structural Entropy~\cite{HY18} and Graph-Based Entropy~\cite{O18} are examples of methods for quantifying the uncertainty associated with model changes. However, these methods do not provide information about the nature or abruptness of the detected changes. Yamanishi and Hirai proposed Descriptive Dimensionality (Ddim), which defines the dimensionality of models when multiple models are fused as a continuous value, and introduced a sign detection method based on tracking changes in Ddim~\cite{YH23}.

The aforementioned sign-detection methods utilize continuous metrics that consider multiple models. However, the individual model dimensions were still treated as discrete values, which is consistent with conventional research. Kyoya and Yamanishi focused specifically on the cluster structure of finite mixture models and introduced the concept of mixture complexity (MC), which considers the overlap and bias between clusters by defining the cluster size as a continuous value~\cite{KY22}. By tracking changes in the cluster size, represented as a continuous value, it is possible to detect the signs of cluster structure changes. Nevertheless, there is room for improvement in terms of sign detection performance because the accurate capture of cluster size relies on the estimation of the number of mixtures in the finite mixture model.

\subsection{Significance of this Work}
The contributions of this study are as follows.

\subsubsection*{1) Proposal of MC fusion: Defining cluster size as a continuous value for finite mixture models with multiple numbers of mixtures}
We propose MC fusion as an extension of the existing concept of MC to address cases in which multiple mixtures are possible in a finite mixture model. MC fusion is a natural extension of mutual information. We present a method for sequentially calculating the MC fusion from time-series data when the distribution is unknown. By incorporating the model fusion method used in Ddim, MC fusion overcomes the limitations of MC, which fails to capture the cluster structure accurately owing to the estimation of the number of mixtures. Unlike Ddim, MC-fusion fusion considers the bias of cluster ratios and the overlap between clusters for each model, providing a more appropriate evaluation of the structure in a continuous manner.

\subsubsection*{2) Empirical demonstration of the effectiveness of the cluster structure change sign detection method using MC fusion}
We propose a cluster structure change sign detection method that focuses on the transition of MC fusion and empirically demonstrate its effectiveness by comparing it with existing methods. In the artificial data analysis, we generated cluster structure changes using Gaussian mixture models with four patterns: splitting, merging, disappearance, and emergence of clusters. We performed sign detection and compared the transition of the MC fusion with other existing methods, demonstrating the effectiveness of our approach by evaluating the scores related to the speed and reliability of detection. Furthermore, we demonstrated the practical application of MC fusion by effectively capturing cluster structure changes in real-world datasets, including COVID-19 infection   and residential power consumption data.

\section{Preliminaries}
In this study, we focus on a scenario in which we observe a dataset $\bm{x}_t$ at each time $t$, and the cluster structure of $\bm{x}_t$ gradually changes over time. Each dataset $\bm{x}_t$ comprises $N$ data points with dimension $d$, which can be represented as $\bm{x}_t = (x_{t,1}, \ldots, x_{t,N})^\top \in \mathbb{R}^{N \times d}$. Our goal was to effectively capture the changes in the cluster structure of the datasets using finite mixture models for clustering.

In this section, we introduce two existing methods that evaluate cluster structures using continuous values: Mixture Complexity (MC)~\cite{KY22}, which continuously quantifies cluster sizes by considering cluster overlap and bias, and Descriptive Dimensionality (Ddim)~\cite{YH23}, which defines model dimensionality as a continuous value. We also discuss the challenges that arise when applying these methods to cluster structural change sign detection.

\subsection{Mixture Complexity}
Let us consider the cluster size of the finite mixture model. The distribution $f$ of a finite mixture model can be expressed as
\begin{align*}
    f(x) = \sum_{i=1}^k \pi_i g(x; \mu_i),
\end{align*}
where $k$ represents the number of mixture components; $\{\pi_i\}_{i=1}^k$ are the parameters representing the probabilities of the data belonging to cluster $i$; and $\{\mu_i\}_{i=1}^k$ are the parameters of distribution $g$. In addition to the observed variable $X$ generated from distribution $f$, we introduce a latent variable $Z \in \{1, \ldots, k\}$ to indicate the cluster from which the observed variable $X$ is generated. The distribution of the latent variable is denoted by $p(Z)$ and the conditional distribution of the observed variable is denoted by $p(X | Z)$:
\begin{align*}
    p(Z = i) = \pi_i, \quad p(X | Z=i) = g(X; \mu_i).
\end{align*}
To capture the cluster structure of the finite mixture model $f$, we consider the mutual information between $Z$ and $X$. The mutual information $I(Z; X)$ can be expressed as
\begin{align*}
    I(Z;X) = H(Z) - H(Z| X),
\end{align*}
where $H(Z)$ represents the entropy of the latent variable $Z$ and $H(Z| X)$ represents the conditional entropy of $Z$. Denoting the posterior probabilities by $\gamma_i(X) = p(Z=i |  X)$, they are defined as follows:
\begin{align*}
    H(Z) &= -\sum_{i=1}^k \pi_i \log \pi_i, \\
    H(Z |  X) &= -E_X\left[\sum_{i=1}^k \gamma_i(X) \log \gamma_i(X)\right].
\end{align*}
Mutual information $I(Z; X)$ represents information about the latent variable contained in the observed data. Therefore, $\exp(I(Z;X))$ can be interpreted as the number of latent variables distinguished by the observed variables and can be considered a continuous measure of cluster size.

However, calculating $I(Z; X)$ is challenging even when the distribution $f$ is known. Therefore, the Mixture Complexity (MC) of a finite mixture model $f$ is defined as an approximation of $I(Z; X)$ using data $\{x_n\}_{n=1}^N$. Specifically, given the posterior probabilities $\gamma_i(x_n)$ for any $i$ and $n$, MC is defined as
\begin{align*}
    \text{MC}\left(\{\gamma_i(x_n)\}_{i,n}\right)
    = -\sum_{i=1}^k \widetilde{\pi}_i \log \widetilde{\pi}_i + \frac{1}{N} \sum_{n=1}^N \sum_{i=1}^k \gamma_i(x_n) \log \gamma_i(x_n),
\end{align*}
where
\begin{align*}
    \widetilde{\pi}_i = \frac{1}{N} \sum_{n=1}^N \gamma_i(x_n).
\end{align*}

\subsection{Descriptive Dimensionality}
The descriptive dimensionality (Ddim) of the model was considered based on the MDL principle~\cite{R78,Y23}. Ddim measures the dimensionality of a model by approximating the parametric complexity. A $k$-dimensional parametric model $\mathcal{P}_k$ with certain regularity conditions is defined as $\mathrm{Ddim}(\mathcal{P}_k) = k$. In other words, Ddim corresponds to the conventional notion of model dimensionality in terms of the degrees of freedom of the parameters. Ddim can also be defined for models that do not have single parameters, making it a natural extension of the traditional notion of dimensionality.

Let us now consider a model $\mathcal{F}^\odot$ where multiple models $\mathcal{P}_1, \ldots , \mathcal{P}_s$ are stochastically fused. In this case, the Ddim of $\mathcal{F}^\odot$ can be lower-bounded as
\begin{align*}
    \mathrm{Ddim}(\mathcal{F}^\odot) \ge \sum_{i=1}^s p(\mathcal{P}_i) \mathrm{Ddim}(\mathcal{P}_i),
\end{align*}
where $p(\mathcal{P}_i)$ represents the probability associated with each model $\mathcal{P}_i$. This lower bound serves as a pseudo-Ddim for model fusion $\mathcal{F}^\odot$. Pseudo-Ddim provides a continuous evaluation of the dimensionality of the fused model and can be calculated when multiple finite mixture models with different mixture components are fused.

\subsection{Issues with Existing Methods}

Both the mixture complexity (MC) and descriptive dimensionality (Ddim) methods are used to evaluate the cluster structure in continuous values. Techniques have been proposed to detect the early signs of gradual changes in the cluster structure by tracking the transitions of MC or Ddim. Here, we outline the respective issues when applying MC and Ddim to detect changes in the cluster structures in a dataset $\bm x_t = (x_{t,1}, \ldots , x_{t,N})^\top \in \mathbb{R}^{N \times d}$ at each time $t$.

\subsubsection{Detecting Cluster Structure Changes using MC}
The MC method relies on estimating the number of components $\hat k_t$ for each time $t$ and then computing the posterior probabilities $\{\hat \gamma_i (x_{t,n}) \}_{i,n}$ using the estimated parameters of the finite mixture model. The MC value $\mathrm{MC}_t = \mathrm{MC} \left( \{\hat \gamma_i (x_{t,n}) \}_{i,n} \right)$ is then calculated. Gradual changes in the cluster structure can be captured by tracking the continuous-valued $\mathrm{MC}_t$.

However, a significant issue with this approach is that the MC depends on the estimated number of components $\hat k_t$. If $\hat k_t$ is underestimated, the MC method may fail to accurately capture the true cluster sizes, particularly when clusters are split and new clusters emerge. These changes may only be considered as a single cluster until $\hat k_t$ changes, leading to a delay in detecting such changes.

\subsubsection{Detecting Cluster Structure Changes using Ddim}
The Ddim method detects changes in the cluster structures by considering the fusion of multiple finite mixture models. The conditional probability $p(k | \bm x_t)$ is estimated for different possible numbers of components $k$ at time $t$. Instead of directly using Ddim, the quantity $\sum_k p(k | \bm x_t)\cdot k$ is computed and denoted by $\mathrm{Ddim}_t$, which can be viewed as a continuous model selection measure. Gradual structural changes can be captured by tracking the changes in $\mathrm{Ddim}_t$ or comparing them with the discrete number of components $\hat k_t$ selected using conventional methods.

However, a limitation of this approach is that it does not capture the structural differences caused by the overlap or bias within each model with a specific number of components $k$. It treats the dimensionality of each model simply as $k$, disregarding variations in the structure due to overlap or bias. Consequently, when gradual changes occur owing to the appearance or disappearance of clusters, it becomes challenging to   capture these changes smoothly using $\mathrm{Ddim}_t$. The transitions in $\mathrm{Ddim}_t$ may exhibit sudden changes similar to those observed in $\hat k_t$, delaying the detection of gradual structural changes.

\section{MC fusion}

In this section, we propose MC fusion as an extension of MC for finite mixture models with multiple possible components.

We attempt to improve the MC by employing the idea of model fusion, as used in Ddim~\cite{YH23} and Structural Entropy~\cite{HY18}. In model fusion, the assumption is made that multiple finite mixture models with different numbers of mixtures are mixed together. Let $\mathcal{K} = \{ k_1, \ldots, k_s \}$ be the set of possible numbers of components, and let $p(K = k)$ denote the probability that the data are generated from a finite mixture model with $k \in \mathcal{K}$ components. In this case, the distribution $f$ can be expressed as:
\begin{align*}
    f(x) &= \sum_{k \in \mathcal{K}} p(K=k) \cdot f_k(x) \\
    &= \sum_{k \in \mathcal{K}} p(K=k) \sum_{i=1}^k \pi_{k,i} g(x; \mu_{k,i}),
\end{align*}
where $f_k$ represents the distribution of a finite mixture model with $k$ components and $\{ \pi_{k,i} \}_{i=1}^k, \{ \mu_{k,i} \}_{i=1}^k$ are the parameters of the distribution $f_k$.

In MC, we fix a single number of components $\hat k$ and compute
\begin{align*}
    I(Z;X | K=\hat k) = H(Z | K=\hat k) - H(Z | X , K=\hat k),
\end{align*}
where $Z$ denotes a latent variable. Under the assumption of model fusion, we can extend this as follows and compute the MC fusion value:
\begin{align*}
    I(Z;X | K) = \sum_{k \in \mathcal{K}} p(K=k) \cdot I(Z;X | K=k).
\end{align*}

When the distribution is unknown, we can compute the MC fusion by estimating $I(Z;X | K=k)$ for each $k \in \mathcal{K}$ as in MC, and estimate $p(K=k)$ using an appropriate method. This calculation method is described in the following section.

\section{Cluster Structure Change Sign Detection Method Using MC fusion}

In this section, we demonstrate how to calculate MC fusion at each time point from the time-series data and describe the application of MC fusion to the detection of predictive signs of cluster structure changes.

\subsection{Normalized Maximum Likelihood (NML) Code Length}

First, we focused on the normalized maximum likelihood (NML) code for the model selection of each data point. The NML code length represents the optimal code length based on Shtarkov's minimax regret~\cite{S87} and provides model selection consistency. Therefore, minimizing the length of the NML code is suitable for accurately estimating the underlying structure. The NML code length for a finite mixture model with $k$ components is as follows.

Let $x^N = x_1, \ldots, x_N$ be the data sequence of observed variables with length $N$. Let $z_n \in \{1, \ldots, k\}$ be the latent variable corresponding to $x_n$ and let $z^N = z_1, \ldots, z_N$.
The NML distribution for the complete variable $(x^N, z^N)$ is given by
\begin{align*}
    p_{\mathrm{NML}} \left( x^N, z^N ; k \right) = \frac{p(x^N, z^N; \hat \theta (x^N, z^N), k)}{\mathcal{C}_N(k)} ,
\end{align*}
and the NML code length for the complete variable $(x^N, z^N)$ is given by
\begin{align*}
    L_{\mathrm{NML}} \left( x^N, z^N ; k \right) &= - \log p_\mathrm{NML}  \left( x^N, z^N ; k \right)  \\
    &= - \log p(x^N, z^N; \hat \theta (x^N, z^N), k) + \log \mathcal{C}_N(k) ,
\end{align*}
where $\hat{\theta}(x^N, z^N)$ is the maximum likelihood estimate of the parameters. In addition, $\mathcal{C}_N(k)$ is the parametric complexity of the finite mixture model with $k$ components and is defined as
\begin{align*}
    \mathcal{C}_N(k) = \int p(\tilde{x}^N, \tilde{z}^N; \hat{\theta}(\tilde{x}^N, \tilde{z}^N), k) \, \mathrm{d}\tilde{x}^N \mathrm{d}\tilde{z}^N,
\end{align*}
where the integral is taken over $\tilde{x}^N$ and $\tilde{z}^N$. The term $\mathcal{C}_N(k)$ can be computed efficiently using the method proposed by Hirai and Yamanishi \cite{HY13,HY19}.

\subsection{MC fusion}

Similar to the calculations in Ddim \cite{YH23}, we can estimate $p(K_t = k)$ for each time step $t$ in the time-series data by considering the NML code lengths and model transition probabilities.

Let $\bm x^T = \bm x_1, \ldots, \bm x_T$ be the time series of the observed data. Here, $\bm x_t = x_{t,1}, \ldots, x_{t,N}$ represents the data observed at time $t$. For each time step $t$, we estimate the latent variable $\bm z_t$ corresponding to $\bm x_t$ using the EM algorithm and define the posterior probability $p(k | \bm x_t, \bm z_t)$ of $k$ given the complete variable $(\bm x_t, \bm z_t)$ as follows:
\begin{align}
    p(k | \bm x_t, \bm z_t) &= \frac{\left(p_\mathrm{NML}(\bm x_t, \bm z_t ; k) \cdot p(k | k_{t-1})\right)^\beta}{\sum_{k'}\left(p_\mathrm{NML}(\bm x_t, \bm z_t ; k') \cdot p(k' |  k_{t-1})\right)^\beta}  \nonumber\\
    &= \frac{\exp\left(-\beta\left(L_\mathrm{NML}(\bm x_t, \bm z_t ; k) - \log p(k | k_{t-1})\right)\right)}{\sum_{k'}\exp\left(-\beta\left(L_\mathrm{NML}(\bm x_t, \bm z_t ; k') - \log p(k' | k_{t-1})\right)\right)},
    \label{p_k}
\end{align}
where $k_{t-1}$ is the estimated mixture number at time $t-1$ and $\alpha$ is a parameter defined as follows:
\begin{align*}
    p(k | k_{t-1}) = \begin{cases}
    1 - \alpha & \text{if $k = k_{t-1}$ and $k \neq 1, k_\mathrm{max}$,} \\
    1 - \alpha / 2 & \text{if $k = k_{t-1}$ and $k = 1, k_\mathrm{max}$,} \\
    \alpha / 2 & \text{if $k = k_{t-1} \pm 1$.}
    \end{cases}
\end{align*}
Parameter $\beta$ is the temperature parameter. In our experiments, we set $\beta = 1 / \sqrt{N}$ following the calculation of Ddim.

Using the above estimations, we can obtain the estimate of $p(K=k)$ as given in (\ref{p_k}), and $I(Z;X |  K=k)$ can be calculated using MC with the parameters estimated for the mixture number $k$. Thus, we can compute the MC fusion as follows:
\begin{align*}
    I(Z;X | K) = \sum_{k \in \mathcal{K}} p(K=k) \cdot I(Z;X | K=k).
\end{align*}

\subsection{Cluster Structural Change Sign Detection}

We propose a method for detecting cluster structural changes based on transitions in MC fusion values over time. By computing the MC fusion at each time step $t$ using the observed data $\bm x^T = {\bm x_1, \ldots, \bm x_T}$, we obtain the MC fusion value $\mathrm{MC\text{-}fusion}(t)$ which provides a continuous evaluation of the cluster structure for the data $\bm x_t$. Visualizing the values of $\mathrm{MC\text{-}fusion}(t)$ over time allowed us to observe the changes in the cluster structure.

To detect cluster structural changes, we propose the following method based on the transitions of MC fusion values.

Let $W$ be the parameter representing the width of the window and $\delta$ be the threshold parameter. We denote the MC fusion value at time $t$ by $y_t$. We raise an alert for cluster structural change if the following condition is satisfied:
\begin{align*}
\left| \mathrm{median}(y_{t-2W+1}, \ldots , y_{t-W}) - \mathrm{median}(y_{t-W+1}, \ldots , y_{t}) \right|   > \delta .
\end{align*}
This condition compares the medians of the MC fusion values within two consecutive windows of width $W$. If the difference between these medians exceeds the threshold $\delta$, the alert is raised, indicating a significant change in the cluster structure. To ensure robust detection, we adopted the median MC fusion values within the window instead of the mean.

The choice of parameters $W$ and $\delta$ depends on the specific application and   desired sensitivity to cluster structural changes. A smaller window width $W$ captures more localized changes, whereas a larger window width provides a broader perspective of the overall changes. Similarly, a smaller threshold $\delta$ increases the sensitivity of detecting smaller changes, whereas a larger threshold focuses on more pronounced changes.

\section{Experiments}

In this section, we present the results of experiments conducted on both artificial and real-world datasets to evaluate the performance of MC fusion in detecting cluster structural changes.

\subsection{Analysis of Artificial Data}

\subsubsection{Datasets}

We generated two types of datasets with gradually changing cluster structures. These datasets contained $T=100$ time steps and $N=1000$ data points, respectively. We considered four patterns of structural changes by examining both the forward and reverse directions. For these datasets, we computed $\text{MC-fusion}(t)$ while selecting the number of mixture components $\hat{k}_t$ using SDMS~\cite{HY12} at each time step $t$. We visualize the transitions of $\text{MC-fusion}(t)$ and compare them with the transitions of other indicators. Furthermore, we compared the performance of sign detection algorithms based on the changes detected by these indicators in terms of detection speed and accuracy.

The first dataset, called the ``moving overlap dataset,'' consists of data where the overlap between clusters gradually changes. Specifically, for each $t$, we generate data $\{ x_{t,n} \}_{n=1}^{1000}$ that follow a 3D Gaussian mixture distribution, as follows:
\begin{align*}
    x_{t,n} \sim \begin{cases}
\mathcal{N} \left( x | \mu = [0,0,0]^\top  \right) & (1 \le n \le 333) ,\\
\mathcal{N} \left( x | \mu = [10,0,0]^\top \right) & (334 \le n \le 666),\\
\mathcal{N} \left( x | \mu = [10 + \alpha(t),0,0]^\top  \right) & (667 \le n \le 1000) ,
\end{cases}
\end{align*}
where the variance of all components is $\Sigma = I_3$ and
\begin{align*}
    \alpha(t) = \begin{cases}
    0 & (1 \le t \le 25),\\
    0.12(t-26) & (26 \le t \le 75),\\
    6 & (76 \le t \le 100).
    \end{cases}
\end{align*}

\begin{figure}
  \centering
  \begin{subfigure}[b]{0.3\columnwidth}
    \includegraphics[width=\textwidth]{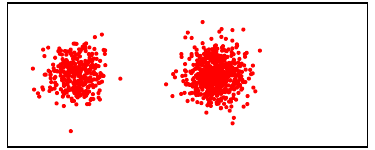}
    \caption{$t=1$}
  \end{subfigure}
  \hfill
  \begin{subfigure}[b]{0.3\columnwidth}
    \includegraphics[width=\textwidth]{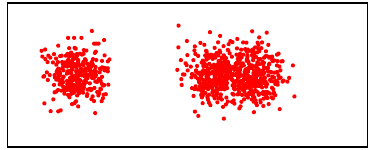}
    \caption{$t=50$}
  \end{subfigure}
  \hfill
  \begin{subfigure}[b]{0.3\columnwidth}
    \includegraphics[width=\textwidth]{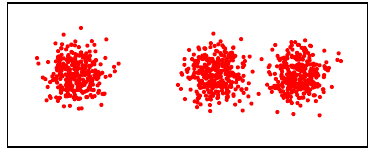}
    \caption{$t=76$}
  \end{subfigure}
  \caption{Plot of data in 2D for the moving overlap dataset at $t=1, 50, 76$.}
  \label{mc_experiment_data1}
\end{figure}

The moving overlap dataset exhibits a gradually changing overlap between the clusters during the transition periods at $t=26, \ldots, 75$. Fig.~\ref{mc_experiment_data1} shows a two-dimensional (2D) visualization of the dataset. In the forward direction ($t=1 \to 100$), the right cluster splits, increasing the number of clusters from two to three. Conversely, in the reverse direction ($t=100 \to 1$), the two right-wing clusters merge, resulting in a decrease in the number of clusters from three to two.

The second dataset, called the ``moving imbalance dataset,'' consists of data where the bias in cluster ratios gradually changes. Specifically, for each $t$, we generate data $\{ x_{t,n} \}_{n=1}^{1000}$ that follow a 3D Gaussian mixture distribution, as follows:
\begin{align*}
    x_{t,n} \sim \begin{cases}
\mathcal{N} \left( x | \mu = [0,0,0]^\top \right) & (1 \le n \le 250) ,\\
\mathcal{N} \left( x | \mu = [10,0,0]^\top  \right) & (251 \le n \le 500),\\
\mathcal{N} \left( x | \mu = [20,0,0]^\top  \right) & (501 \le n \le 750 + \alpha(t)) ,\\
\mathcal{N} \left( x | \mu = [30,0,0]^\top  \right) & (751 + \alpha(t) \le n \le 1000) ,
\end{cases}
\end{align*}
where the variance of all components is $\Sigma = I_3$ and
\begin{align*}
    \alpha(t) = \begin{cases}
    0 & (1 \le t \le 25),\\
    5(t-26) & (26 \le t \le 75),\\
    250 & (76 \le t \le 100).
    \end{cases}
\end{align*}

\begin{figure}
  \centering
  \begin{subfigure}[b]{0.3\columnwidth}
    \includegraphics[width=\textwidth]{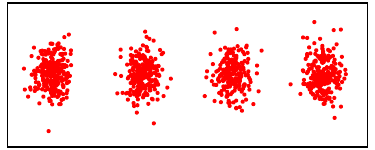}
    \caption{$t=1$}
  \end{subfigure}
  \hfill
  \begin{subfigure}[b]{0.3\columnwidth}
    \includegraphics[width=\textwidth]{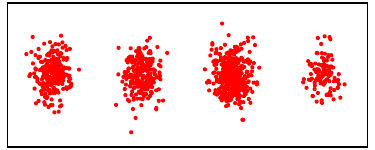}
    \caption{$t=55$}
  \end{subfigure}
  \hfill
  \begin{subfigure}[b]{0.3\columnwidth}
    \includegraphics[width=\textwidth]{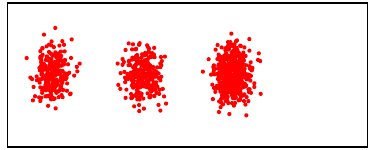}
    \caption{$t=76$}
  \end{subfigure}
  \caption{Plot of data in 2D for the moving imbalance dataset at $t=1, 55, 76$.}
  \label{mc_experiment_data2}
\end{figure}

The moving imbalance dataset exhibits a gradually changing bias in the cluster ratios during the transition periods at $t=26, \ldots, 75$. Fig.~\ref{mc_experiment_data2} shows a two-dimensional (2D) visualization of the dataset. In the forward direction ($t=1 \to 100$), the rightmost cluster disappeared, resulting in a decrease in the number of clusters from four to three. Conversely, in the reverse direction ($t=100 \to 1$), a cluster appeared on the right side, leading to an increase in the number of clusters from three to four.

\subsubsection{Methods for Comparison}

The following existing methods were compared with the MC fusion:

\begin{enumerate}
    \item Mixture Complexity (MC)~\cite{KY22}: MC defines the cluster size as a continuous value by considering the overlap and bias between clusters in a finite mixture model with a selected number of components. It evaluates a single selected model and does not assume model fusion.

    \item Descriptive Dimensionality (Ddim)~\cite{YH23}: Ddim defines the model's dimensionality as a continuous value. It assumes model fusion.

    \item SDMS~\cite{HY12}: SDMS is an extension of DMS to the sequential setting. It sequentially selects the number of components based on the MDL principle and outputs a sequence of models.

    \item Fixed Share algorithm (FS)~\cite{HW98}: FS treats experts in an aggregating algorithm as candidate models and updates the weights of experts sequentially to track the best expert.
\end{enumerate}

The SDMS and FS are conventional methods that iteratively select the number of models discretely. On the other hand, MC and Ddim are continuous evaluation methods that aim to detect structural changes early.

\subsubsection{Evaluation Metric}

For each method, the early detection of structural changes and the reliability of detection have a trade-off relationship. Therefore, to evaluate the performance of each method in terms of both detection speed and accuracy while considering this trade-off, the Area Under Curve (AUC) of the benefit-false alarm rate (FAR) curve was defined as an evaluation metric.

The benefit metric evaluates the speed of detection, and is defined as follows:
\begin{align*}
    \mathrm{benefit} = \begin{cases}
        1 - (\hat t - t^\ast) / U & (t^\ast \le \hat t  < t^\ast + U), \\
        0 & \text{otherwise},
    \end{cases}
\end{align*}
where $\hat t$ is the time when the first alarm is raised during the true transition period and $t^\ast$ is the starting point of the true change. $U$ is a parameter, and in this experiment, it was set to $U=25$, which is half of the transition period.

The FAR measures the ratio of alarms raised at time points not included in the transition period to the total number of time points not included in the transition period. This quantifies the reliability of the detection.

Let $y_t$ denote the values calculated at time $t$ for $\text{MC-fusion} (t)$, $\mathrm{MC}_t$, $\log (\mathrm{Ddim}_t)$, and the selected number of components $\hat k_t$ by SDMS and FS. Because the true change (increase or decrease in cluster number) was known in this experiment, an alarm was raised when
\begin{align*}
    \mathrm{median}(y_{t-4}, \ldots , y_{t}) - \mathrm{median}(y_{t-9}, \ldots , y_{t-5})  > \delta
\end{align*}
for an increasing number of clusters, and when
\begin{align*}
    \mathrm{median}(y_{t-4}, \ldots , y_{t}) - \mathrm{median}(y_{t-9}, \ldots , y_{t-5})  < - \delta
\end{align*}
as the number of clusters decreased. 
To evaluate the performance of the sign detection algorithm, the benefit and FAR are computed for various threshold values $\delta$. By plotting the benefit-FAR curve, the performance of the algorithm was assessed based on the AUC.

In addition, $\text{delay} = \hat{t} - t^*$ was compared for each method, particularly when the threshold was set to $\delta = 0.01$. The delay indicates how quickly the method detects a change compared to the true starting point of the change.

\subsubsection{Results}
The results of the performance comparison among the methods based on AUC and delay for the four types of structural changes are presented in Table \ref{table_compare}. Based on the results presented in Table \ref{table_compare}, we interpret the differences between the methods and visualize the transitions for each type of structural change. However, because the same results were obtained from SDMS and FS, we compared the estimated number of components $k$ by using MC fusion, MC, and Ddim.

\begin{table*}
\centering
\caption{AUC and Delay scores for each method on the four patterns of structural changes.}
\label{table_compare}
\begin{tabular}{c|cccccccc}
          & \multicolumn{4}{c}{Moving overlap dataset}            & \multicolumn{4}{c}{Moving imbalance dataset}                      \\
Method    & \multicolumn{2}{c}{Split} & \multicolumn{2}{c}{Merge} & \multicolumn{2}{c}{Disappearance} & \multicolumn{2}{c}{Emergence} \\
          & AUC          & Delay      & AUC         & Delay       & AUC              & Delay          & AUC          & Delay          \\ \hline \hline
MC fusion & 0.712        & 28         & 0.995       & 10          & 0.928            & 13             & 1.0          & 2              \\
MC        & 0.590        & 34         & 0.994       & 14          & 1.0              & 14             & 1.0          & 2              \\
Ddim      & 0.638        & 28         & 0.985       & 14          & 0.709            & 36             & 1.0          & 2              \\
SDMS      & 0.5          & 34         & 0.660       & 19          & 0.5              & 51             & 1.0          & 2              \\
FS        & 0.5          & 34         & 0.660       & 19          & 0.5              & 51             & 1.0          & 2             
\end{tabular}
\end{table*}

\begin{figure}
  \centering
  \begin{subfigure}[b]{\columnwidth}
    \includegraphics[width=\textwidth]{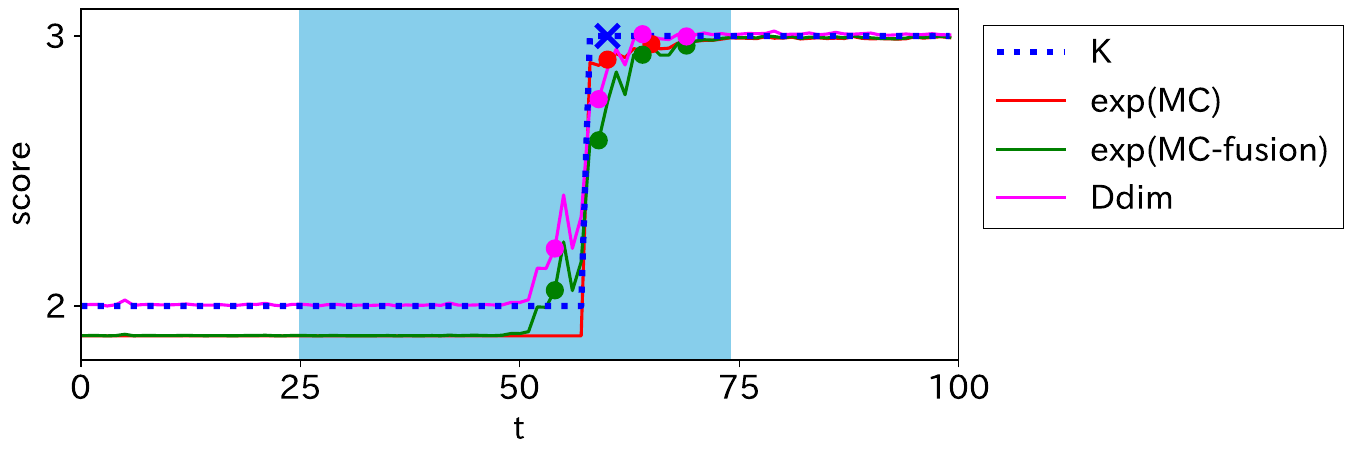}
    \caption{$t = 1 \to 100$}
  \end{subfigure}

  \vspace{0.5cm}

  \begin{subfigure}[b]{\columnwidth}
    \includegraphics[width=\textwidth]{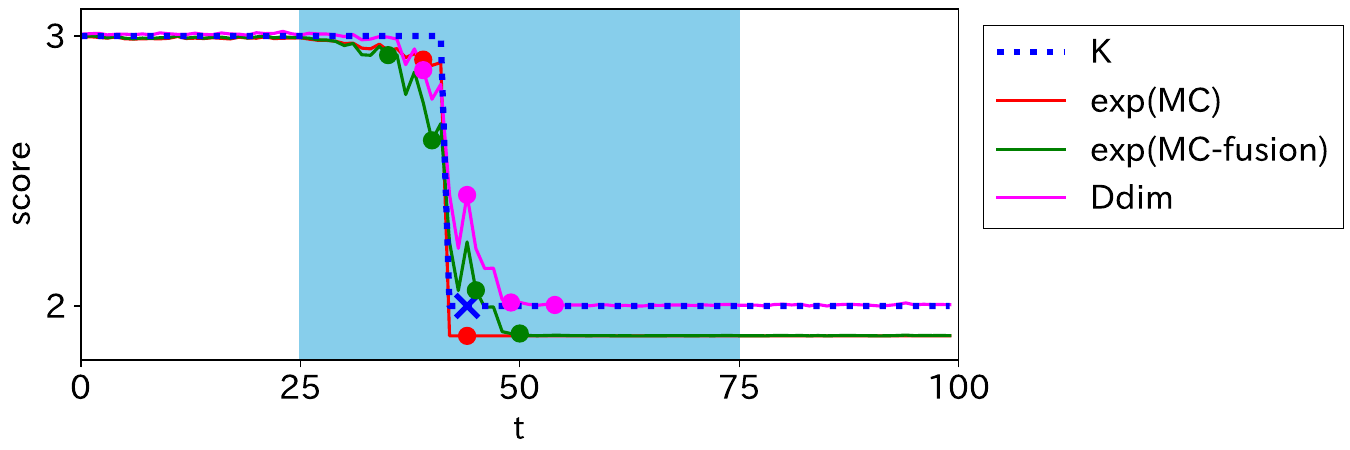}
    \caption{$t = 100 \to 1$}
  \end{subfigure}

  \caption{Estimated values of $k$, $\exp(\mathrm{MC})$, $\exp(\mathrm{MC\text{-}fusion})$, and $\mathrm{Ddim}$ for each time $t$ in the moving overlap dataset. The light blue area represents the transitional period of change, and markers indicate the points where each method raised change alerts.}
  \label{mc_experiment_result1}
\end{figure}

For the moving overlap dataset, Fig. ~\ref{mc_experiment_result1} illustrates the transitions of $k$, $\exp (\mathrm{MC})$, $\exp(\text{MC-fusion})$, $\mathrm{Ddim}$ estimated at time $t$. Focusing on the $t = 1 \to 100$ direction, which corresponds to cluster splitting, MC does not show a noticeable increase until $k$ changes from two to three, indicating that MC is not effective for early detection. By contrast, both MC fusion and Ddim start to increase before the change in $k$, enabling early detection. The delay for $k$ and MC was 34, whereas it was 28 for MC fusion and Ddim. Moreover, the AUC score for MC was lower than those for MC fusion and Ddim. In the $t = 100 \to 1$ direction, which corresponds to cluster merging, all three methods (MC, MC fusion, and Ddim) show a decrease in values from the stage where $k$ is estimated to be three, indicating the early detection of a decrease in the number of clusters. MC fusion achieved the best scores in terms of both the AUC and delay, whereas MC and Ddim performed similarly.

\begin{figure}
  \centering
  \begin{subfigure}[b]{\columnwidth}
    \includegraphics[width=\textwidth]{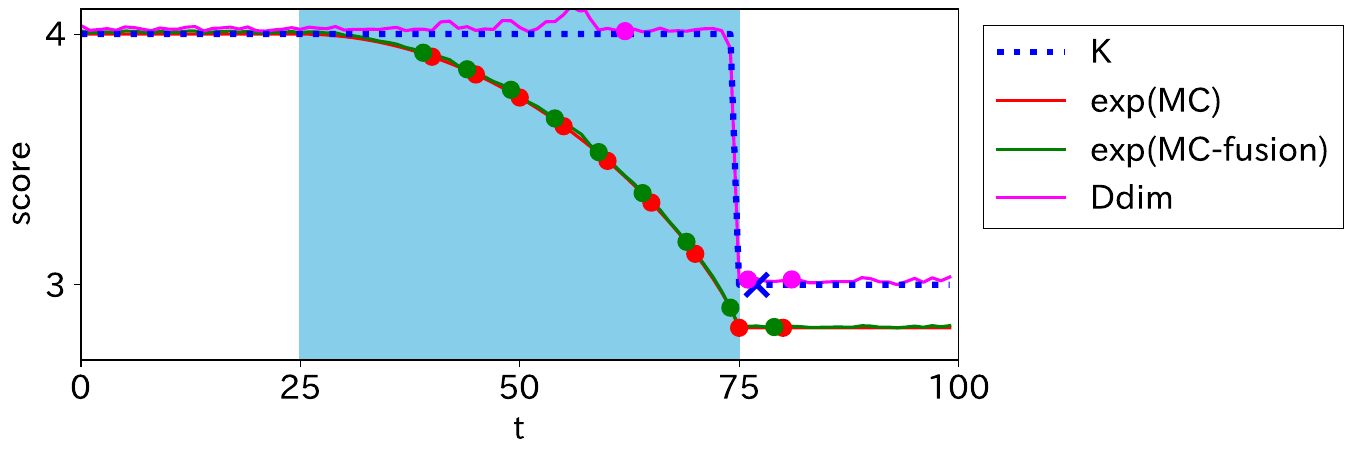}
    \caption{$t = 1 \to 100$}
  \end{subfigure}

  \vspace{0.5cm}

  \begin{subfigure}[b]{\columnwidth}
    \includegraphics[width=\textwidth]{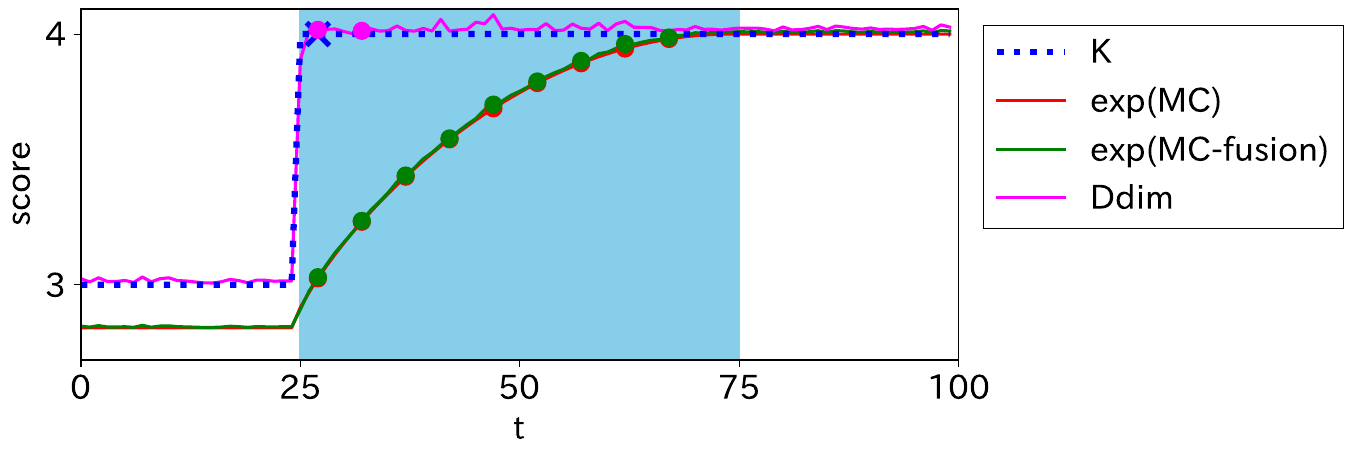}
    \caption{$t = 100 \to 1$}
  \end{subfigure}

  \caption{Estimated values of $k$, $\exp(\mathrm{MC})$, $\exp(\mathrm{MC\text{-}fusion})$, and $\mathrm{Ddim}$ for each time $t$ in the moving imbalance dataset. The light blue area represents the transitional period of change, and markers indicate the points where each method raised change alerts.}
  \label{mc_experiment_result2}
\end{figure}

For the moving imbalanced dataset, Fig. ~\ref{mc_experiment_result2} shows the transitions of $k$, $\exp (\mathrm{MC})$, $\exp(\text{MC-fusion})$, $\mathrm{Ddim}$ estimated at each time $t$. Focusing on the $t = 1 \to 100$ direction, which corresponds to cluster disappearance, MC fusion, which is similar to MC, captures the gradual decrease in cluster size and detects early signs of change. However, the estimated value of $k$ remains unchanged until the transition period is complete, and Ddim follows a trajectory similar to that of $k$, making it unable to detect the change early. The delay for Ddim was more than 20 units higher than those for MC fusion and MC, and the AUC score was particularly low. In the $t = 100 \to 1$ direction, which corresponds to cluster emergence, both MC and MC fusion smoothly captured the changes in cluster structure, and early detection was possible with all methods.

\subsection{Analysis of COVID-19 infection data}

\subsubsection{Dataset}
We analyzed a dataset of COVID-19 cases and deaths\footnote{\url{https://github.com/CSSEGISandData/COVID-19}}. The dataset included daily data on new infections and deaths in various countries. We calculated the number of infected individuals, individuals with immunity, and deaths at each time point in each country. We also calculated the ratios of these quantities to the population in each country. The dataset covers the period from January 22, 2020, to October 17, 2022, for 182 countries.

\subsubsection{Results}
\begin{figure}
  \centering
  \includegraphics[width=\columnwidth]{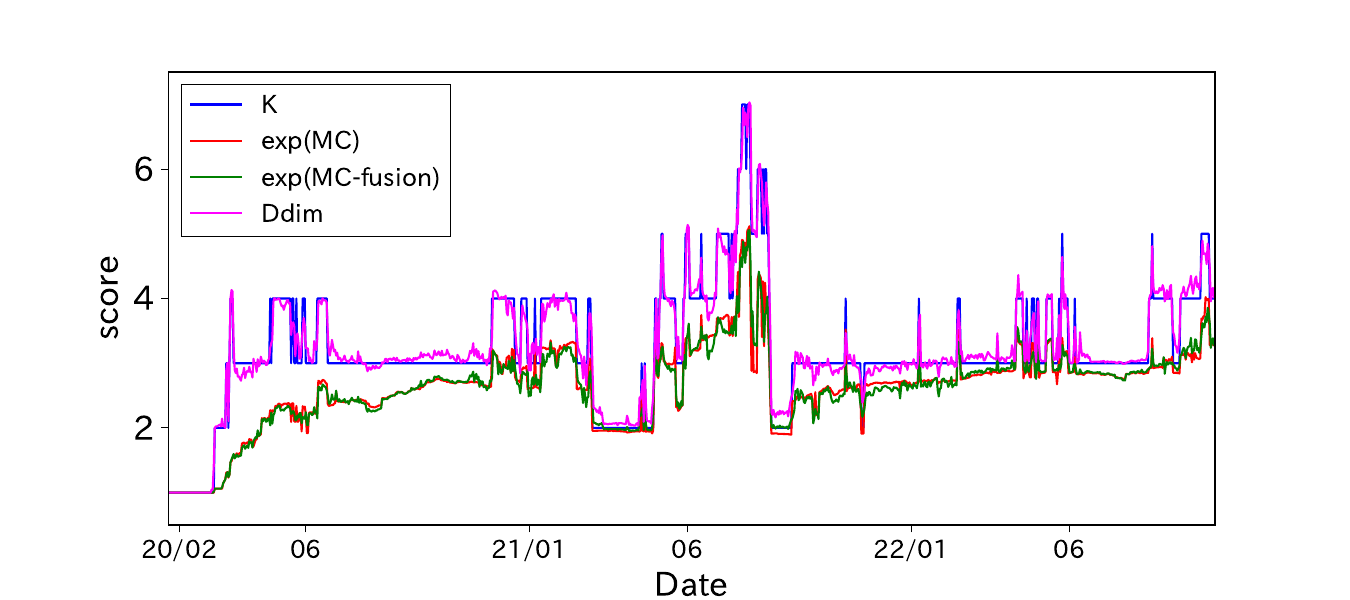}
  \caption{Estimated values of $k$, $\exp(\mathrm{MC})$, $\exp(\mathrm{MC\text{-}fusion})$, and $\mathrm{Ddim}$ for each day $t$ in the COVID-19 infection data (throughout the entire period).}
  \label{mc_experiment_covid_all}
\end{figure}

Fig.~\ref{mc_experiment_covid_all} illustrates the transitions in $k$, $\exp (\mathrm{MC})$, $\exp(\text{MC-fusion})$, and $\mathrm{Ddim}$ over time for the COVID-19 data set. These clusters represent the patterns of infection in each country. Before the outbreak, all countries had values close to zero for infected individuals (I), individuals with immunity (R), and deaths (D). However, over time, countries have been divided into those experiencing outbreaks and those that have not, with several outbreak patterns observed. These changes were likely influenced by differences in infection control measures and the movement of people between countries.

\begin{figure}
  \centering
  \begin{subfigure}[b]{0.8\columnwidth}
    \includegraphics[width=\textwidth]{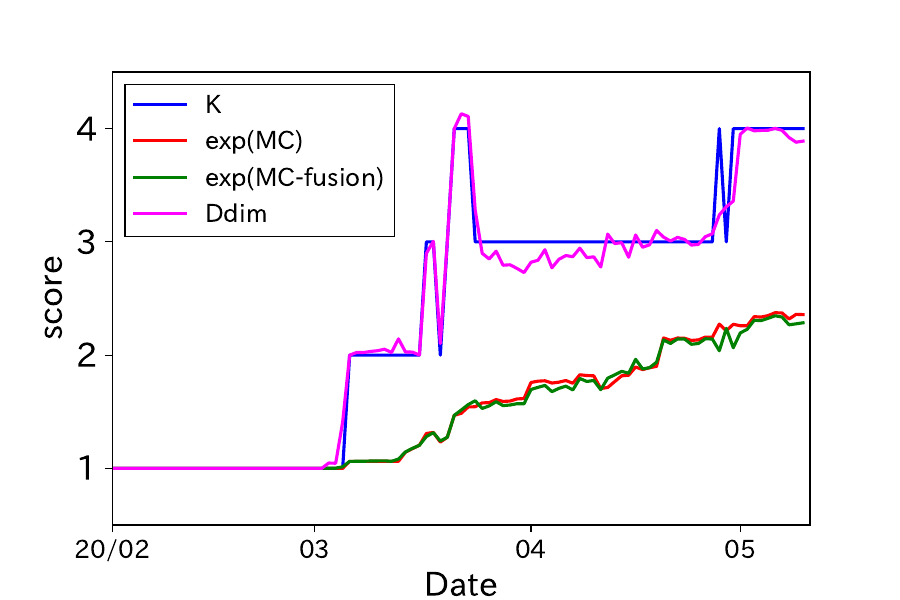}
    \caption{From February 1, 2020, to May 10, 2020.}
    \label{covid_2to5}
  \end{subfigure}
  \vspace{0.5cm}
  \begin{subfigure}[b]{0.8\columnwidth}
    \includegraphics[width=\textwidth]{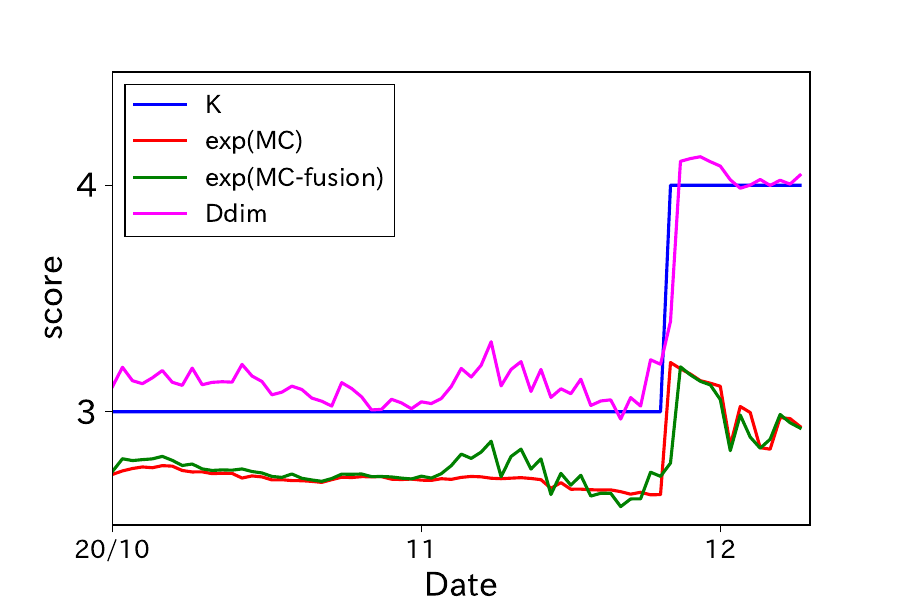}
    \caption{From October 1, 2020, to December 9, 2020.}
    \label{covid_10to12}
  \end{subfigure}
  \caption{Partial excerpts of the analysis of COVID-19 infection data, specifically focusing on the early and middle stages of the infection.}
  \label{mc_experiment_covid_partial}
\end{figure}

Owing to numerous factors influencing infection patterns and their impacts more than a year after the start of the outbreak, we specifically examined two periods in 2020 when the estimated mixture number $k$ changed. Fig.~\ref{mc_experiment_covid_partial} presents a subset of the analysis. First, to gain a detailed understanding of the changes that occurred around April 2020, Table \ref{tab_change1} shows the transition of the cluster means and the number of countries belonging to each cluster. In Table \ref{tab_change1}, we observe three clusters: pre-outbreak countries (c1), countries where the outbreak began (c2), and countries that experienced an explosive outbreak (c3). While the number of clusters remained constant, the number of countries with outbreaks gradually increased. Fig.~\ref{covid_2to5} illustrates that despite the constant estimation of $k$, MC and MC fusion successfully captured gradual changes in the cluster structure.

Moving on to the changes that occurred around November 2020, Table \ref{tab_change2} presents the transition of cluster means and the number of countries belonging to each cluster. From Table \ref{tab_change2}, we observe that the expanding outbreak cluster c3 is split into two clusters: c3 and c4. Cluster c3 indicates a decrease in the number of infected individuals and an increase in recovered individuals, suggesting that the outbreak subsided in these countries. By contrast, cluster c4 showed an increase in the number of infected individuals, implying an ongoing outbreak in these countries. Fig.~\ref{covid_10to12} demonstrates that $k$ and MC capture this cluster split on November 26th, while the value of MC fusion starts to deviate upward in early November, enabling the early detection of an increase in the number of clusters on November 5th when using a detection threshold of $\delta=0.01$.

\begin{table*}
    \centering
    \caption{Transition of cluster means and the number of countries belonging to each cluster around April 2020. Note that the unit of cluster means is $10^{-4}$.}
    \label{tab_change1}
    \begin{subtable}[b]{\columnwidth}
        \centering
        \caption{April 1, 2020.}
        \begin{tabular}{c|ccc}
          cat. & c1 & c2 & c3 \\
          \hline \hline
          I & 0.34 & 4.7 & 28 \\
          R & 0.055 & 1.1 & 8.0 \\
          D & 0.007 & 0.21 & 1.7 \\
          \hline
          \# & 147 & 26 & 9
        \end{tabular}
    \end{subtable}
    \\
    \vspace{0.5cm}
    \begin{subtable}[b]{\columnwidth}
        \centering
        \caption{April 15, 2020.}
        \begin{tabular}{c|ccc}
          cat. & c1 & c2 & c3 \\
          \hline \hline
          I & 0.44 & 6.3 & 25 \\
          R & 0.36 & 6.3 & 49 \\
          D & 0.026 & 0.80 & 3.1 \\
          \hline
          \# & 140 & 37 & 5
        \end{tabular}
    \end{subtable}
    \\
    \vspace{0.5cm}
    \begin{subtable}[b]{\columnwidth}
        \centering
        \caption{April 29, 2020.}
        \begin{tabular}{c|ccc}
          cat. & c1 & c2 & c3 \\
          \hline \hline
          I & 0.24 & 4.5 & 9.7 \\
          R & 0.49 & 6.2 & 37 \\
          D & 0.029 & 0.29 & 3.3 \\
          \hline
          \# & 121 & 44 & 17
        \end{tabular}
    \end{subtable}
\end{table*}

\begin{table*}
    \centering
    \caption{Transition of cluster means and the number of countries belonging to each cluster around November 2020. Note that the unit of cluster means is $10^{-4}$.}
    \label{tab_change2}
    \begin{subtable}[b]{\columnwidth}
        \centering
        \caption{November 10, 2020.}
        \begin{tabular}{c|ccc}
          cat. & c1 & c2 & c3 \\
          \hline \hline
          I & 0.32 & 21 & 49 \\
          R & 5.8 & 63 & 210 \\
          D & 0.13 & 1.4 & 5.7 \\
          \hline
          \# & 73 & 68 & 41
        \end{tabular}
    \end{subtable}
    \\
    \vspace{0.5cm}
    \begin{subtable}[b]{\columnwidth}
        \centering
        \caption{November 20, 2020.}
        \begin{tabular}{c|ccc}
          cat. & c1 & c2 & c3 \\
          \hline \hline
          I & 0.54 & 26 & 44 \\
          R & 7.1 & 85 & 250 \\
          D & 0.15 & 1.8 & 6.4 \\
          \hline
          \# & 79 & 65 & 38
        \end{tabular}
    \end{subtable}
    \\
    \vspace{0.5cm}
    \begin{subtable}[b]{\columnwidth}
        \centering
        \caption{December 1, 2020.}
        \begin{tabular}{c|cccc}
          cat. & c1 & c2 & c3 & c4\\
          \hline \hline
          I & 0.54 & 13 & 40 & 70 \\
          R & 6.4 & 79 & 290 & 190 \\
          D & 0.14 & 1.5 & 6.1 & 6.8 \\
          \hline
          \# & 75 & 55 & 34 & 18
        \end{tabular}
    \end{subtable}
\end{table*}

\subsection{Analysis of Electric Power Consumption Data}

\subsubsection{Dataset}
We analyzed a dataset of residential electric power consumption obtained from the UCI machine learning repository ~\cite{DG17}. The dataset covers the period from December 16, 2006, to November 26, 2010, for three distribution networks in Tetouan, northern Morocco. The data were recorded at 15-minute intervals each day, denoted by $t = 1, \ldots , 96$. Each observation $x_{t,n}$ represents electric power consumption in three categories (kitchen, laundry, and air conditioning) at a specific time interval $t$ and location $n$. The dataset contained 4326 observations.

\subsubsection{Results}
\begin{figure}
  \centering
  \includegraphics[width=\columnwidth]{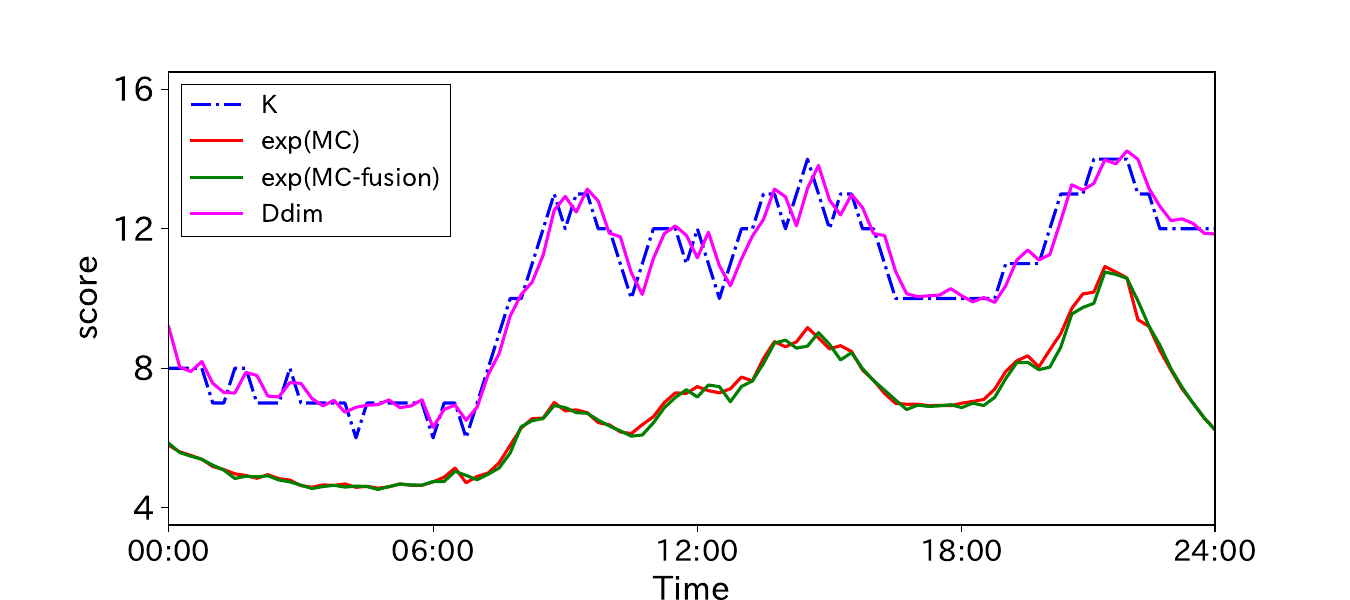}
  \caption{Estimation of $k$, $\exp(\mathrm{MC})$, $\exp(\mathrm{MC}\text{-fusion})$, and $\mathrm{Ddim}$ for residential power consumption data at each time interval $t$.}
  \label{mc_experiment_house}
\end{figure}

Fig.~\ref{mc_experiment_house} illustrates the transitions of $k$, $\exp (\mathrm{MC})$, $\exp(\text{MC-fusion})$, and $\mathrm{Ddim}$ over time for the electric power consumption dataset. The clusters in this context represent the patterns of electric power consumption in the RHs. For instance, an increase in the number of clusters around 7 a.m. indicates that many people wake up at that time and start their daily activities, resulting in diverse power consumption patterns. By focusing on the cluster structure, it is possible to capture latent changes in the data. In particular, both MC and MC fusion were capable of detecting changes in the cluster structure at approximately 7 a.m. and 7 p.m. Furthermore, both methods demonstrated the ability to capture essential changes without being overly sensitive to minor and less significant variations in the data.

\subsection{Discussion}

The experiments conducted in this study provide valuable insights into the effectiveness of the cluster structure change sign detection method using MC fusion. The key findings and conclusions based on the experiments with both artificial and real datasets are as follows:

\subsubsection*{1) Performance comparison with existing methods}
MC fusion outperformed the existing methods, MC and Ddim, in terms of early detection of structural changes. MC fusion was particularly effective in detecting cluster splitting, which MC failed to capture. Ddim, however, struggled to detect cluster disappearance, whereas MC fusion captured changes in cluster bias.

\subsubsection*{2) Consistency of results}
The trends observed in the experiments using artificial data were consistent with those observed in the experiments using real data. In the analysis of COVID-19 infection data, both MC fusion and MC successfully captured changes in cluster bias as the number of affected countries increased during the early stages of the pandemic. MC fusion demonstrated the ability to detect early signs of cluster splitting, whereas MC alone was unable to do so. However, in some cases, such as the analysis of residential electric power consumption data, MC fusion exhibits a behavior similar to that of   MC for longer durations.

\subsubsection*{3) Effectiveness of MC fusion}
Compared to Ddim, MC fusion proved to be more effective in capturing essential changes without being overly sensitive to minor variations in the data. This is because of the consideration of cluster overlap and bias in the MC fusion. Compared to MC, MC fusion exhibited similar behavior for longer durations but was capable of detecting structural changes, such as cluster splitting, at an earlier stage.

\section{Conclusion}

In this study, we present MC fusion as an extension of Mixture Complexity (MC) for finite mixture models. MC fusion incorporates the bias of cluster proportions and the overlap between clusters by fusing multiple models with different mixture numbers. We demonstrated the effectiveness of MC fusion in detecting signs of cluster structure changes using both artificial and real datasets. MC fusion outperformed existing methods, such as MC and Ddim, in terms of the early detection and capture of changes in the cluster structure.

In future research, it would be valuable to further investigate and refine the methods and criteria for raising alarms regarding the signs of cluster structure changes. This includes exploring the optimal threshold and considering the parameter $\beta$ used in the estimation of the probability $p(K=k)$, which determines the mixture number $k$. Optimizing the selection of $\beta$ as a hyperparameter can enhance the MC fusion performance. In addition, there are numerous potential applications to explore, such as the analysis of communities in network data as clusters.

Overall, MC fusion offers a promising approach for detecting and monitoring changes in cluster structures and provides valuable insights and applications in various domains.

\section*{Statements and Declarations}
\textbf{Funding: }This work was partially supported by JST KAKENHI 19H01114.

$ $

\noindent\textbf{Competing Interests: }The authors declare that they have no conficts of interest.

\appendix
\section{Implementation Code}
The code used in our experiments can be accessed at the GitHub repository \url{https://github.com/uraken38/MC-fusion}. Please refer to README for instructions.
The experiments were conducted on a server with the following specifications: an Intel Core i5 processor (1.6GHz), 8GB of RAM, and a 234GB solid-state drive. Programming was performed using Python version 3.7.13.

\end{document}